# MemSeg: A semi-supervised method for image surface defect detection using differences and commonalities

Minghui Yang*, Peng Wu, Jing Liu, and Hui Feng

*Xidian University*

**Abstract**

Under the semi-supervised framework, we propose an end-to-end memory-based segmentation network (MemSeg) to detect surface defects on industrial products. Considering the small intra-class variance of products in the same production line, from the perspective of **differences** and **commonalities**, MemSeg introduces artificially simulated abnormal samples and memory samples to assist the learning of the network. In the training phase, MemSeg explicitly learns the potential differences between normal and simulated abnormal images to obtain a robust classification hyperplane. At the same time, inspired by the mechanism of human memory, MemSeg uses a memory pool to store the general patterns of normal samples. By comparing the similarities and differences between input samples and memory samples in the memory pool to give effective guesses about abnormal regions; In the inference phase, MemSeg directly determines the abnormal regions of the input image in an end-to-end manner. Through experimental validation, MemSeg achieves the state-of-the-art (SOTA) performance on MVTec AD datasets with AUC scores of **99.56%** and **98.84%** at the image-level and pixel-level, respectively. In addition, MemSeg also has a significant advantage in inference speed benefiting from the end-to-end and straightforward network structure, which better meets the real-time requirement in industrial scenarios.

## 1. Introduction

The detection of product surface anomalies in industrial scenarios is crucial to the development of industrial intelligence. Surface defect detection is a problem of locating abnormal regions in images, such as scratches and smudges. But in practical applications, anomaly detection by traditional supervised learning is more difficult due to the low probability of abnormal samples and the diverse forms of anomalies. Therefore, the methods based on semi-supervised techniques for surface defect detection have more significant advantages in practical applications, which require only normal samples in the training phase.

Based on semi-supervised techniques, most image surface defect detection models attempt to explore the general patterns of normal samples efficiently. For example, reconstruction models based on autoencoder (AE) [2,3] or generative adversarial network (GAN) [4,5,6,7] aim to reconstruct normal images with minimal error and locate anomalies based on the reconstruction error. But due to the powerful generalization ability of CNN [8,9], abnormal regions may also be reconstructed correctly in the inference stage, which clearly violates the basic assumptions of the reconstruction models. Recently, the embedding-based methods [10,11,12,13] have shown better anomaly detection performance than reconstruction-based methods. The fundamental principle of the embedding-based methods is the feature matching between the test samples and the normal samples. Although such models require little time consumption in the training phase, they need to perform complex feature matching operations in the inference phase, which incurs excessive computational costs for the inference of the model. Besides, such models are not trained using anomaly-specific datasets and directly use pre-trained parameters for feature extraction and anomaly detection, which is not sufficiently adaptable to the anomaly detection task.

---

\* Corresponding author.
   *E-mail address:* areylng@163.com (M. Yang).

Given the shortcomings of existing methods, we propose an end-to-end memory-based segmentation network (MemSeg) in this paper to accomplish the defect detection of product surface. Instead of reconstructing the input images, our model determines the abnormal regions in the images end-to-end. Also, our model does not entirely rely on the pre-trained model for feature extraction, which alleviates the problem of inconsistent distribution between the source and target domains. The design of MemSeg is based on the observation of the **small intra-class variance** of products in the same industrial production line, we believe that artificially creating intra-class differences and preserving intra-class commonalities can help the model achieve better defect detection performance. These two types of information, differences and commonalities, can provide a more direct orientation to the model learning and help the model generalize normal patterns more comprehensively and distinguish non-normal patterns more precisely.

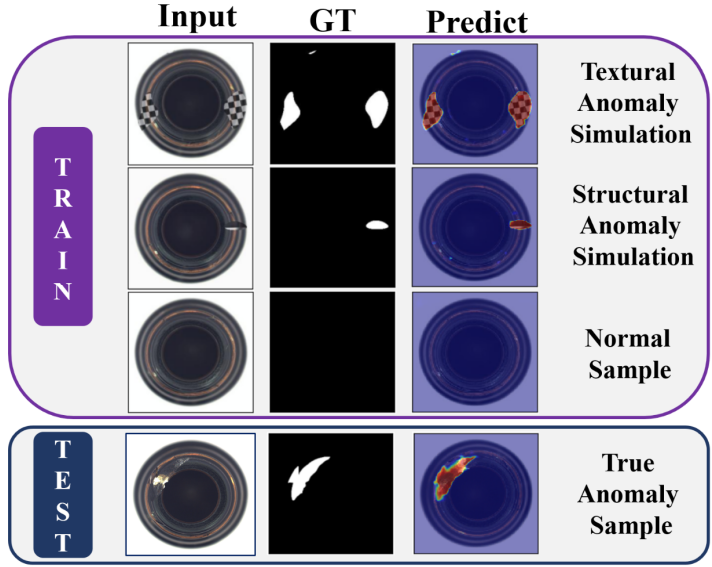

Fig. 1. The data usage during the training and testing phases. The images are taken from the MVTec AD dataset [1].

Specifically, from the perspective of **differences**, similar to self-supervised learning, MemSeg introduces artificially simulated anomalies during the training phase to make the model consciously discriminate normal from non-normal while does not require the simulated anomalies are consistent with those in the real scenarios, which alleviates the deficiency that semi-supervised learning can only use normal samples and allows the model to obtain a more robust decision boundary. MemSeg uses normal and simulated abnormal images to finish the model training and directly judges the abnormal regions of the input images without any auxiliary tasks in the inference phase. Fig. 1 shows our data usage during the training and inference phases.

Meanwhile, from the perspective of **commonalities**, MemSeg introduces a memory pool to record the general patterns of normal samples. In the training and inference phases of the model, we compare the similarities and differences between the input samples and the memory samples in the memory pool to provide more effective information for the localization of abnormal regions. In addition, in order to coordinate the information from the memory pool and the input image more effectively, MemSeg introduces a multi-scale feature fusion module and a novelty spatial attention module, which substantially improves the performance of the model.

With the target of precise localization of abnormal regions in images, MemSeg achieves state-of-the-art (SOTA) performance with 99.56% and 98.84% AUROC scores in image-level and pixel-level, respectively, on the MVTec AD dataset [1], which contains 75 different forms of anomalies in real scenarios. Meanwhile, MemSeg also has outstanding performances on other datasets. More precise but faster, MemSeg can process 31.34 images per second in the inference phase using an NVIDIA RTX 3090 GPU, which better meets the real-time requirements in industrial production.

To summarize, the main contributions of this paper are:
- We propose a well-designed anomaly simulation strategy for self-supervised learning of the model, which integrates the three aspects of target foreground, textural and structural anomalies.
- We propose a memory module with a more efficient feature matching algorithm, and innovatively introduce the memory information of normal patterns in the U-Net structure to assist the model learning.
- Through the two points above, and combining the multi-scale feature fusion module and the spatial attention module, we effectively simplify semi-supervised anomaly detection into an end-to-end semantic segmentation task, making semi-supervised image surface defect detection more flexible.
- Through extensive experimental validation, MemSeg has high accuracy in surface defect detection and localization tasks while better meeting the real-time requirements of industrial scenarios.

## 2. Related Works

### 2.1. Reconstruction-Based Methods

One of the traditional methods for image anomaly detection is reconstruction-based methods. Most reconstruction-based methods use autoencoder (AE) [2,3] or generative adversarial network (GAN) [4,5,6,7] to train a network to reconstruct the input images. The basic assumption of the reconstruction models is that the model can reconstruct normal images with a small error and reconstruct abnormal images with a large error. However, in practical applications, the learning ability of neural networks is too strong [8,9], so the abnormal regions in the image may also be reconstructed well, and anomaly discrimination based on the reconstruction error may be invalid. To reduce the influence of abnormal regions on the reconstruction models, RIAD [15], based on an autoencoder, performs a multi-scale complementary mask operation on the original image and tries to cover the abnormal regions using the mask. Similarly, InTra [16], based on a transformer, uses the image with a masked patch as the input and completes a patch repair task. However, in any case, the reconstruction models are still affected by the abnormal regions in the inference phase because the exact locations of abnormal regions are not clear. MemSeg is still based on an AE but avoids the reconstruction process of the input image and completes the anomaly localization in an end-to-end manner.

### 2.2. Anomaly Simulation-Based Methods

The semi-supervised models of image surface defect detection use only normal samples as training data. For the models to explicitly learn the potential differences between normal and abnormal samples, some works [17,18,19] attempted to generate artificially simulated abnormal samples during training. Specifically, DRAEM [17] superimposes additional texture images as noise onto the normal images to generate abnormal regions, and this type of data augmentation method aims to create textural anomalies. CutPaste [18] and AnoSeg [19] use an augmentation method similar to copy and paste. This kind of method randomly copies a small rectangular area from the input image and randomly pastes it to the image to simulate abnormal samples. By pasting rectangular patches of different sizes, aspect ratios, and rotation angles to create structural anomalies. As a means of data processing, the existing anomaly simulation methods only consider structural anomalies or textural anomalies one-sidedly; at the same time, for some datasets, there is a problem of low simulation efficiency because the target foreground and background in images cannot distinguish well. However, the anomaly simulation strategy used by MemSeg solves these shortcomings.

In addition, despite the introduction of simulated abnormal samples, AnoSeg and DRAEM still need to complete the reconstruction process of the input image; CutPaste only completes the defect detection at the image level, and the defect localization at the pixel level is implemented by GradCAM or Gaussian density estimation. More directly, with the help of our well-designed anomaly simulation strategy, our model does not need the reconstruction of the input image as an auxiliary task for model learning and completes the defect localization at the pixel-level end-to-end.

*2.3. Embedding-Based Methods*

Embedding-based methods [10,11,12,13] usually use a pre-trained network on ImageNet [20] to extract the high-level features of the original images, and the anomaly score is calculated through the distance between the test sample and the normal samples on the features to obtain the abnormal regions. FYD [10] designs a two-stage coarse-to-fine feature alignment network that learns robust feature distributions of normal images; SPADE [12] extends the KNN anomaly detection method to pixel-level and detects anomalies in the images through the pixel-level correspondence between the test image and normal images; PaDiM [13] uses a pre-trained CNN to extract the patch embeddings of the input image and uses the multivariate Gaussian distribution to obtain the probability representation of the normal samples. Due to the simplicity and effectiveness, embedding-based models are widely used, but they usually require a complex features matching process in the inference phase, which greatly limits the inference speed of models. The memory module in our model is still based on the principle of embedding-based methods, but through the design of a more efficient feature matching algorithm, the memory module does not add too much computational cost to the model while guaranteeing the model precision.

## 3. Method

In this section, we demonstrate our novel framework to detect and localize fine-grained anomalies. An overview of MemSeg is shown in Fig. 2. MemSeg uses U-Net [21] as a framework to complete a semantic segmentation task with the help of simulated abnormal samples and memory information in the training phase, and localizes abnormal regions in images end-to-end in the inference phase. MemSeg consists of several essential parts, we will describe these parts in the following order: generation of abnormal samples by the way of artificial simulation (Subsection 3.1), generation of memory information and spatial attention maps (Subsection 3.2), multi-scale feature fusion module (MSFF Module) for the fusion of memory information and high-level features of images (Subsection 3.3), and loss functions (Subsection 3.4).

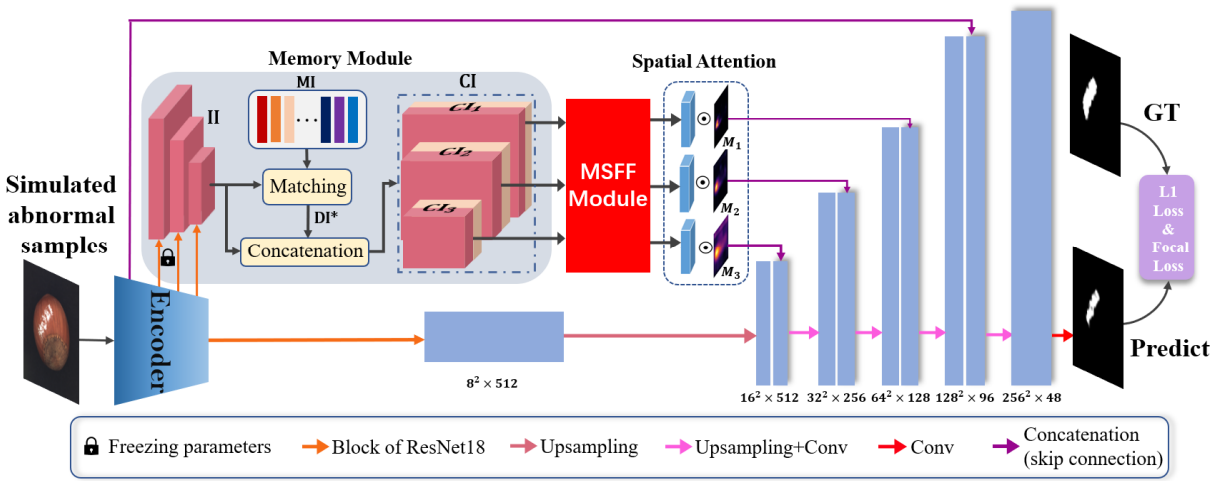

Fig.2. An overview of MemSeg. MemSeg is based on the U-Net architecture and uses a pre-trained ResNet18 [14] as an encoder. From the perspective of differences and commonalities, MemSeg introduces simulated abnormal samples and a memory module to assist the model learning in a more oriented way, and thus accomplishes the semi-supervised surface defect task in an end-to-end manner. At the same time, in order to fully fuse the memory information with the high-level features of the input image, MemSeg introduces a multi-scale feature fusion module (MSFF Module) and a novel spatial attention module, which greatly improves the model precision of anomaly localization.

*3.1. Anomaly Simulation Strategy*

In industrial scenarios, anomalies occur in various forms, and it is impossible to cover all of them when performing data collection, which limits the modeling with the supervised learning methods. However, in the semi-supervised framework, using only normal samples and no comparisons with non-normal samples is not sufficient for the model to learn what are the normal patterns. In this paper, inspired by DRAEM [17], we design a more effective strategy to simulate abnormal samples and introduce them during training to accomplish self-supervised learning. MemSeg summarizes the patterns of normal samples by comparing non-normal patterns to mitigate the drawbacks of semi-supervised learning. As shown in Fig. 3, the anomaly simulation strategy proposed in this paper is mainly divided into three steps.

**In the first step**, a two-dimensional Perlin noise [22] $P$ is generated, then $P$ is binarized by threshold $T$ to obtain the mask $M_P$. The Perlin noise has several random peaks, and $M_P$ generated by it can extract contiguous blocks of regions in the image. At the same time, considering that the proportion of the main body of some industrial components in the acquired image is small, if data augmentation is performed directly without processing, it is easy to generate noise in the background part of the image, which increases the differences between simulated abnormal samples and real abnormal samples on the data distribution, which is not conducive for the model to learn effective discriminative information, so we adopt a foreground enhancement strategy for this type of images. That is, the input image $I$ is binarized to obtain the mask $M_I$, and the noise generated in the binarization process is removed using the open or close operation. After that, the final mask image $M$ is obtained by performing an element-wise product on the two obtained masks.

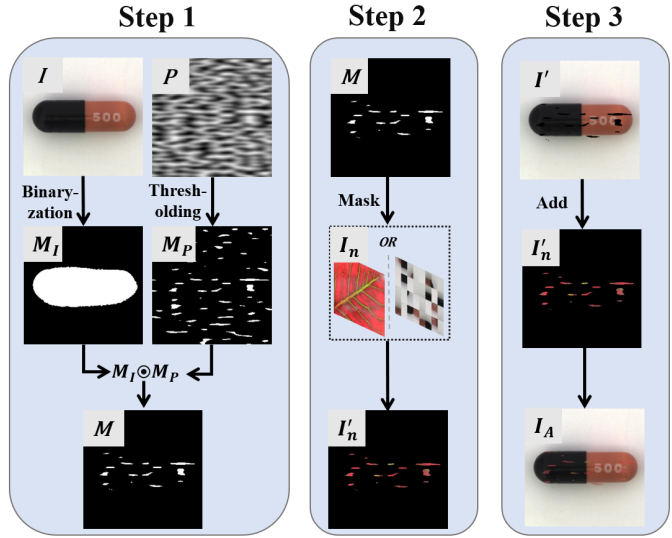

Fig.3. The three steps of our anomaly simulation strategy. In the first step, the mask image $M$ is generated using Perlin noise and target foreground; in the second step, the ROI defined by $M$ in the noise image $I_n$ is extracted to generate the noise foreground image $I'_n$; in the third step, the noise foreground image is superimposed on the original image to obtain the simulated abnormal image $I_A$.

**In the second step**, the mask image $M$ and the noise image $I_n$ perform element-wise product to get the region of interest (ROI) in $I_n$ defined by $M$. Consistent with DRAEM [17], we introduce a transparency factor $\delta$ in this process to balance the fusion of the original image and the noisy image, so that the patterns of simulated anomalies are closer to the real anomalies. Therefore, the noisy foreground image $I'_n$ is generated using the following equation:

$$I'_n = \delta(M \odot I_n) + (1 - \delta)(M \odot I) \tag{1}$$

For the noisy image $I_n$, we want its maximum transparency to be higher to increase the difficulty of model learning and thus improve the robustness of the model. So for $\delta$ in Eq. 1, we will randomly and uniformly sample from [0.15, 1].

**In the third step**, the mask image $M$ is inverted to obtain $\bar{M}$, then the element-wise product is performed on $\bar{M}$ and the original image $I$ to obtain image $I'$, and according to

$$I_A = \bar{M} \odot I + I'_n \tag{2}$$

the data-augmented image $I_A$, namely, the simulated abnormal image, is obtained. $I_A$ takes the original input image $I$ as the background and the ROI in the noise image $I_n$ extracted by mask image $M$ as the foreground.

Among them, the noisy image $I_n$ comes from **two parts**, one part from the DTD texture dataset [23], which aims to simulate textural anomalies, the other part from the input image, which aims to simulate structural anomalies. For the simulation of structural anomalies, we first perform random adjustment of mirror symmetry, rotation, brightness, saturation, and hue on the input image $I$. Then the preliminary processed image is uniformly divided into a 4×8 grid and randomly arranged to obtain the disordered image $I_n$.

With the above anomaly simulation strategy, we obtain simulated abnormal samples from both textural and structural perspectives, and most abnormal regions are generated on the target foreground, which maximizes the similarity between the simulated abnormal samples and the real abnormal samples.

*3.2. Memory Module and Spatial Attention Maps*

**Memory Module.** For humans, our identification of anomalies is predicated on knowing what is normal, and the abnormal regions are obtained by comparing the test image with the normal images in our memory. Inspired by the human learning process and embedding-based methods, we use a small number of normal samples as memory samples and extract high-level features of the memory samples as memory information using a pre-trained encoder (ResNet18 [14]) to assist the learning of MemSeg.

To obtain the memory information, we first randomly select $N$ normal images from the training data as memory samples and input them to the encoder to get features of dimensions $N×64×64×64$, $N×128×32×32$, and $N×256×16×16$ from block 1, block 2, and block 3 of ResNet18, respectively. These features with different resolutions together constitute the memory information $MI$. It needs to be emphasized that in order to ensure the unification of the memory information and the high-level features of the input images, we always freeze the model parameters of block 1, block 2, and block 3 in ResNet18, but the rest of the model is still trainable.

Given an input image in the training or inference phase, as shown in Fig. 2, the encoder also extracts high-level features of the input image to obtain features with dimensions of 64×64×64, 128×32×32, and 256×16×16. These features with different resolutions together constitute the information of the input image $II$. After that, the L2 distance between the $II$ and all the memory information $MI$ is calculated, so $N$ difference information $DI$ between the input image and the memory samples is obtained:

$$DI = \bigcup_{i=1}^{N} \|MI_i - II\|_2 \qquad (3)$$

where $N$ is the number of memory samples. For $N$ difference information, take the minimum sum of all elements in each $DI$ as the standard to obtain the best difference information $DI^*$ between $II$ and $MI$; that is,

$$DI^* = \underset{DI_i \in DI}{\mathrm{argmin}} \sum_{x \in DI_i} x \qquad (4)$$

where $i \in [1, N]$. The best difference information $DI^*$ contains the differences between the input sample and its most similar memory sample, the larger the difference value at a position, the higher the probability that the region of the input image corresponding to that position is abnormal.

Subsequently, the best difference information $DI^*$ completes the concatenation operation with the high-level features of the input image $II$ in the channel dimension to obtain the concatenated information $CI_1$, $CI_2$ and $CI_3$. Finally, the concatenated information will go through the multi-scale feature fusion module for feature fusion, and the fused features flow to the decoder through the skip connection of U-Net.

**Spatial Attention Maps.** It is evident from specific observations and experiments (Subsection 4.6) that the best difference information $DI^*$ has an important influence on the localization of abnormal regions. To make full use of the difference information, we extract three spatial attention maps using $DI^*$, which are used to reinforce the guesses of best difference information on the abnormal regions.

For the features with three different dimensions in $DI^*$, the mean values are calculated in the channel dimension, and three feature maps of size 16×16, 32×32 and 64×64 are obtained, respectively. The 16×16 feature map is directly used as the spatial attention map $M_3$. After $M_3$ is up-sampled, perform the element-wise product operation with the 32×32 feature map to obtain $M_2$; and after $M_2$ is up-sampled, perform the element-wise product operation with the 64×64 feature map to obtain $M_1$. As shown in Fig. 2, spatial attention map $M_1$, $M_2$ and $M_3$ weighted the information which obtained after $CI_1$, $CI_2$ and $CI_3$ are processed by the MSFF, respectively. Mathematically, the formulas for solving $M_1$, $M_2$ and $M_3$ are given as follows:

$$M_3 = \frac{1}{C_3} \sum_{i=1}^{C_3} DI_{3i}^* \qquad (5)$$

$$M_2 = \frac{1}{C_2} \left( \sum_{i=1}^{C_2} DI_{2i}^* \right) \odot M_3^U \qquad (6)$$

$$M_1 = \frac{1}{C_1} \left( \sum_{i=1}^{C_1} DI_{1i}^* \right) \odot M_2^U \qquad (7)$$

where $C_3$ denotes the number of channels of $DI_3^*$; $DI_{3i}^*$ denotes the feature map of channel $i$ in $DI_3^*$; $M_3^U$ and $M_2^U$ denote the feature maps obtained after up-sampling $M_3$ and $M_2$, respectively.

### 3.3. Multi-Scale Feature Fusion Module

With the help of the memory module, we obtain the concatenated information **CI** composed of the input image information $II$ and the best difference information $DI^*$. The direct use of $CI$ has the problem of feature redundancy on the one hand; on the other hand, it increases the computational scale of the model and causes a decrease in the inference speed. Given the success of multi-scale feature fusion in target detection [24,25], an intuitive idea is to fully fuse the visual information and semantic information in the concatenated information $CI$ with the help of the channel attention mechanism and multi-scale feature fusion strategy.

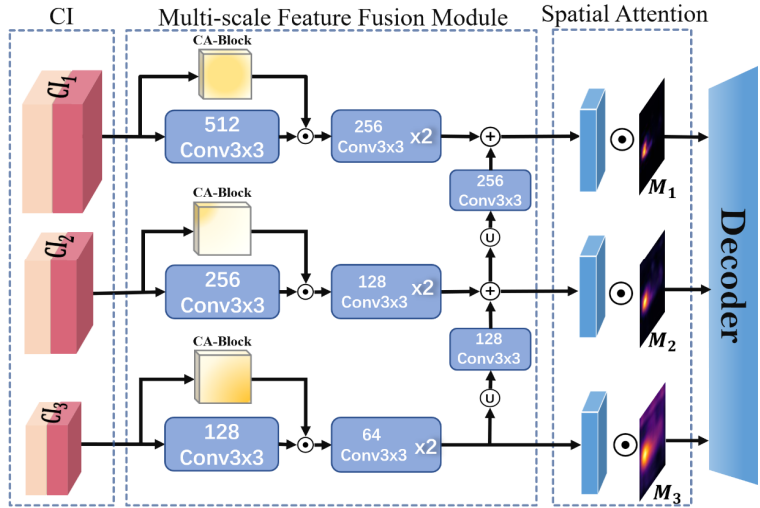

Fig.4. The multi-scale feature fusion module used by MemSeg. Considering that $CI_n$ is a concatenation of two kinds of information in the channel dimension, and comes from different locations of the encoder with different semantic information and visual information, so we use channel attention CA-Block and multi-scale strategy for feature fusion.

Our proposed multi-scale feature fusion module is shown in Fig. 4: the concatenated information $CI_n$ ($n = 1,2,3$) is initially fused by a 3×3 convolutional layer that maintains the number of channels. Meanwhile, considering that $CI_n$ is a simple concatenation of two kinds of information in the channel dimension, we use coordinate attention (CA) [26] to capture the information relationship between channels of $CI_n$. Then, for the features with different dimensions weighted by coordinate attention, we continue perform multi-scale information fusion: the feature maps of different dimensions are firstly aligned in resolution using up-sampling, then aligned in the number of channels using convolution, finally the operation of element-wise add is executed to achieve multi-scale feature fusion. The fused features are weighted by the spatial attention maps $M_n$ ($n = 1,2,3$) obtained in Subsection 3.2 and then fed to the final decoder.

*3.4. Training Constraints*

To ensure that the predicted value of the MemSeg is close to its ground truth, we use L1 loss and focal loss [27] to guarantee the similarity of all pixels in the image space. The segmentation images predicted under the L1 loss constraint retain more edge information compared to the L2 loss. Meanwhile, focal loss alleviates the problem of area imbalance between normal and abnormal regions in images and makes the model focus more on the segmentation of difficult samples to improve the accuracy of abnormal segmentation.

Specifically, we minimize the L1 loss $L_{l1}$ and the focal loss $L_f$ between the ground truth $S$ of the abnormal regions in the simulated image and the predicted value $\hat{S}$ of the model using (8) and (9), respectively.

$$L_{l1} = \|S - \hat{S}\|_1 \tag{8}$$

$$L_f = -\alpha_t(1 - p_t)^\gamma \log(p_t) \tag{9}$$

where $p_t$ is equal to the predicted probability $p$ of the pixel category when the ground truth of the corresponding pixel in $S$ is 1, and $p_t = 1 - p$ when the ground truth of the pixel in $S$ is 0, $\alpha_t$ and $\gamma$ as hyperparameters to control the degree of weighting.

Finally, we combine all these constraints into an objective function, and arrive at the following objective function:

$$L_{all} = \lambda_{l1} L_{l1} + \lambda_f L_f \tag{10}$$

where $\lambda_{l1}$ and $\lambda_f$ are the balancing hyper-parameters. In the training process, our optimization goal is to minimize the objective function defined by Eq. (10).

## 4. Experiments

In this section, we evaluate the performance of MemSeg as well as the functionalities of different components on the semi-supervised anomaly detection datasets: MVTec AD dataset [1], BeanTech AD dataset [28], and a toy dataset.

*4.1. Datasets and Evaluation Metric*

The MVTec AD dataset [1] is mainly aimed to the task of semi-supervised surface anomaly detection. The MVTec AD dataset comprises 15 categories, including 5 different texture categories and 10 different object categories, each category includes about 60 to 400 normal samples for training and a mixture of normal and abnormal images for testing, and the test set contains a variety of realistic anomalies with different textures and scales. The BeanTech dataset [28] has 3 categories of 2540 images, which also only contain normal images in the training set.

For the evaluation metric on the image surface, following the works in [10,11,12,13,17,18,19], we leverage image-level and pixel-level ROC-AUC for performance evaluation.

*4.2. Implementation Details*

MemSeg is based on the AE model, it uses ResNet18 [14] as the encoder. And for the decoder part, corresponding to Fig. 2, the up-sampling layer contains a bilinear interpolation layer and a basic convolution block consisting of a convolution layer, batch-normalization, and a ReLU activation function. The Conv Layer contains two stacked basic convolution blocks; only the last Conv Layer contains one basic convolution block and a 2-channel convolution layer. The training process of MemSeg is carried out with 2700 iterations, the size of the input image is set to 256×256, and the batch size is set to 8, which contains 4 normal samples and 4 simulated abnormal samples. When performing anomaly simulation, most categories have an equal probability of using textural anomaly simulation and structural anomaly simulation. We use a grid search to set hyper-parameters: the learning rate used is set to 0.04; $\gamma$ in the focal loss is set to 4; $\lambda_{l1}$ and $\lambda_f$ in the objective function is set to 0.6 and 0.4, respectively.

For most of the categories in both datasets, we randomly select 30 memory samples in the training set to generate the memory information, but since the orientation of the screws in the MVTec AD dataset is randomly arranged, we increased the number of their memory samples for better feature matching; the sample size of toothbrushes in the training set is too small, so we use only 10 memory samples while ensuring adequate training samples. MemSeg obtains the anomaly score for each pixel in the image in an end-to-end manner, and the mean of the scores of the top 100 most abnormal pixel points in the image is used as the anomaly score at the image-level.

*4.3. Comparison with Existing Methods*

In this subsection, we compare MemSeg with different methods. The AUC scores of different methods are listed in Tables 1 and 2. We can see that our method outperforms most existing methods, which demonstrates the effectiveness of our method.

**Table 1.** The comparison between our method and different methods on the MVTec AD dataset in terms of ROC-AUC % with the format of (**Image-level, Pixel-level**).

| | Category | SPADE [12] | PaDiM [13] | DRAEM [17] | CutPaste [18] | P-SVDD [29] | P-SVDD-C [30] | Ours |
|---|---|---|---|---|---|---|---|---|
| Texture | carpet | (-,97.5) | (-,98.9) | (97.0,95.5) | (92.9,92.6) | (93.1,98.3) | (94.4,92.9) | (**99.6**,**99.2**) |
| | grid | (-,93.7) | (-,94.9) | (99.9,**99.7**) | (94.6,96.2) | (99.9,97.5) | (95.6,97.2) | (**100**,99.3) |
| | leather | (-,97.6) | (-,99.1) | (100,98.6) | (90.9,97.4) | (100,99.5) | (96.1,98.2) | (**100**,**99.7**) |
| | tile | (-,87.4) | (-,91.2) | (99.6,99.2) | (97.8,91.4) | (93.4,90.5) | (93.5,91.9) | (**100**,**99.5**) |
| | wood | (-,85.5) | (-,93.6) | (99.1,96.4) | (96.5,90.8) | (98.6,95.5) | (98.0,92.1) | (**99.6**,**98.0**) |
| | average | (-,92.3) | (-,95.6) | (99.1,97.9) | (94.5,93.7) | (97.0,96.3) | (95.5,94.5) | (**99.8**,**99.1**) |
| Object | bottle | (-,98.4) | (-,98.1) | (99.2,99.1) | (98.6,98.1) | (98.3,97.6) | (99.5,98.6) | (**100**,**99.3**) |
| | cable | (-,97.2) | (-,95.8) | (91.8,94.7) | (90.3,96.8) | (80.6,90) | (97.8,**97.6**) | (**98.2**,97.4) |
| | capsule | (-,99.0) | (-,98.3) | (98.5,94.3) | (76.7,95.8) | (96.2,97.4) | (88.7,96.3) | (**100**,**99.3**) |
| | hazelnut | (-,99.1) | (-,97.7) | (100,**99.7**) | (92.0,97.5) | (97.3,97.3) | (97.9,98.2) | (**100**,98.8) |
| | metal nut | (-,98.1) | (-,96.7) | (98.7,**99.5**) | (94.0,98.0) | (99.3,93.1) | (96.5,98.1) | (**100**,99.3) |
| | pill | (-,96.5) | (-,94.7) | (98.9,97.6) | (86.1,95.1) | (92.4,95.7) | (91.9,92.4) | (**99.0**,**99.5**) |
| | screw | (-,**98.9**) | (-,97.4) | (93.9,97.6) | (81.3,95.7) | (86.3,96.7) | (83.3,95.3) | (97.8,98.0) |
| | toothbrush | (-,97.9) | (-,98.7) | (100,98.1) | (100,98.1) | (98.3,98.1) | (95.6,96.0) | (**100**,**99.4**) |
| | transistor | (-,94.1) | (-,97.2) | (93.1,90.9) | (91.5,97) | (95.5,93.0) | (92.1,93.5) | (**99.2**,**97.3**) |
| | zipper | (-,96.5) | (-,98.2) | (100,98.8) | (97.9,95.1) | (99.4,**99.3**) | (95.9,96.0) | (**100**,98.8) |
| | average | (-,97.57) | (-,97.3) | (97.4,97.0) | (90.8,96.7) | (94.3,95.8) | (93.9,96.2) | (**99.4**,**98.7**) |
| Average | | (85.5,96.0) | (95.3,96.7) | (98.0,97.3) | (95.2,96.0) | (92.1,95.7) | (94.4,95.6) | (**99.56**,**98.84**) |

**Table 2.** The comparison between our method and different methods on the BeanTech AD dataset in terms of ROC-AUC % with the format of (**Image-level, Pixel-level**).

| Category | PatchCore [11] | SPADE [12] | PaDiM [13] | P-SVDD [29] | Ours |
|---|---|---|---|---|---|
| 01 | (90.9,95.5) | (91.4,97.3) | (**99.8**,97.0) | (95.7,91.6) | (98.7,**98.9**) |
| 02 | (79.3,94.7) | (71.4,94.4) | (82.0,96.0) | (72.1,93.6) | (**87.0**,**96.2**) |
| 03 | (99.8,**99.3**) | (**99.9**,99.1) | (99.4,98.8) | (82.1,91.0) | (99.4,96.3) |
| Mean | (90.0,96.5) | (87.6,96.9) | (93.7,**97.3**) | (83.3,92.1) | (**95.0**,97.1) |

For the MVTec AD dataset, at the image-level, the category with the worst anomaly detection effect is the screw. On the one hand, the reason is that our model relies more on the alignment of detection targets in space, but the orientation of screws in the dataset is randomly arranged, which makes it difficult to generate effective difference information; on the other hand, the reason is that some abnormal regions of screws in the test set are small and difficult to distinguish, and the model is prone to misclassification, which is also reflected in other models. At the pixel-level, the category with the worst anomaly detection effect is the transistor, part of the reason is that when the transistor in the circuit is missing or the direction is shifted, it is difficult for our model to give a precise anomaly localization.

For the BeanTech AD dataset, although the anomaly localization of MemSeg at the pixel-level is not as good as PaDiM (97.1% vs 97.3%), the AUC score of MemSeg is better than all the models in the experiment at the image-level, which indicates that our model still has an accurate anomaly detection capability for other datasets.

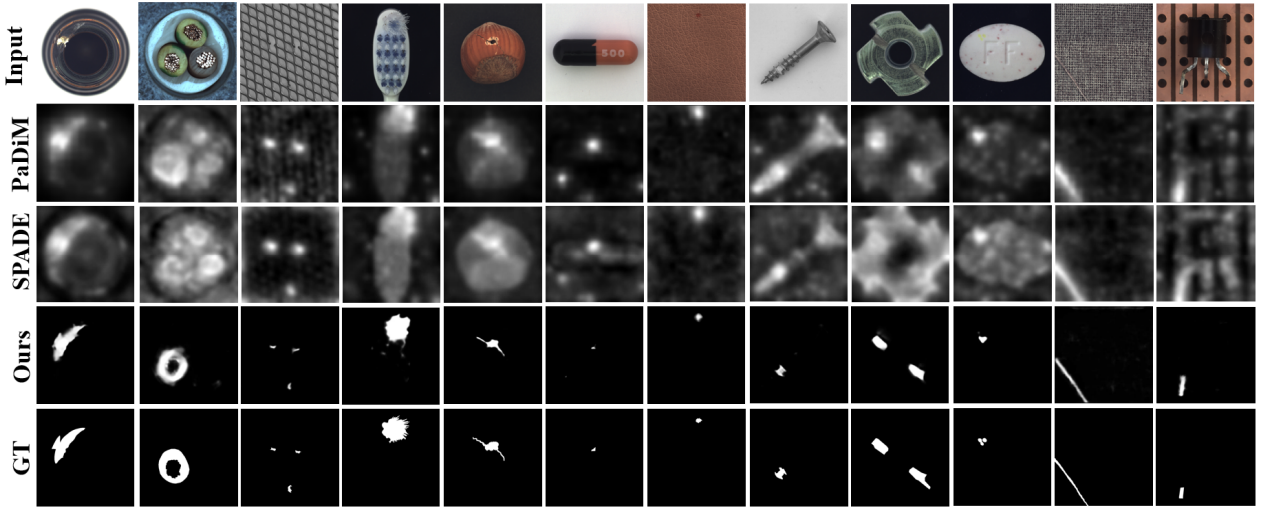

Fig.5. Comparison of MemSeg with PaDiM [13] and SPADE [12] for anomaly localization on the MVTec AD dataset (before thresholding). Our model has a more precise judgment of the abnormal regions.

Meanwhile, as shown in Fig. 5, the anomaly localization of MemSeg at the pixel-level is closer to the ground truth, and the boundary between the normal and abnormal regions is more precise, which benefits from the end-to-end learning approach adopted by MemSeg and the training of the model is directly guided by the pixel-level ground truth of simulated abnormal samples.

*4.4. Impact of Anomaly Simulation Strategy*

To evaluate the effectiveness of the proposed anomaly simulation strategy for image defect detection, we remove textural anomaly simulation, structural anomaly simulation, and foreground enhancement strategy in training, respectively, and compare the three cases with our complete strategy. We report the AUC scores at the image-level and pixel-level for each of the above four experiments in Table 3. The AUC scores decrease when either component of the anomaly simulation strategy is removed, which verifies that our anomaly simulation strategy is not only theoretically interpretable, but also has an excellent performance in experimental validation.

Table 3. Evaluating the components of our anomaly simulation strategy on the MVTec AD dataset. The AUC scores are reported for different strategies.

|             | w/o Texture | w/o Structure | w/o Foreground | All   |
|-------------|-------------|---------------|----------------|-------|
| Image-level | 98.80       | 98.77         | 99.34          | **99.56** |
| Pixel-level | 97.31       | 98.09         | 98.40          | **98.84** |

Now, to evaluate the role of simulated abnormal samples more fully, we also want to know the data distribution of simulated and real abnormal samples after the training of the model, so we visualize the output of the encoder of simulated abnormal samples, the real abnormal samples in the test set and the normal samples using t-SNE [31]. As shown in Fig. 6, for most categories, there is some overlapping in the spatial distribution of simulated abnormal samples and the real abnormal samples, while the abnormal samples are separated from the normal samples, which proves the validity of our anomaly simulation strategy.

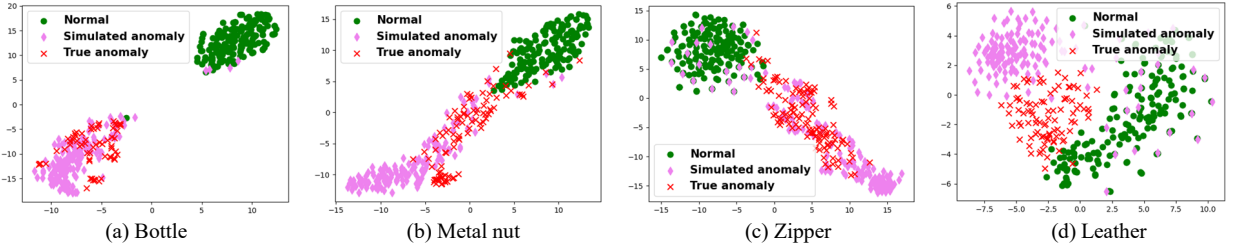

Fig.6. Separability display of normal samples, simulated abnormal samples and real abnormal samples in the MVTec AD dataset.

It is important to emphasize that the features we visualized are generated only at the bottleneck structure of U-Net. Although the separability of some features in some categories is not strong in two-dimensional space, our model can still be corrected in the decoder part using information from the skip connection. Besides, MemSeg does **not** completely require the distribution of simulated abnormal samples to be the same as real abnormal samples. MemSeg is based on the semi-supervised learning framework, the reason we introduce the simulated abnormal samples during training is simply to make the model explicitly learn the difference between normal and non-normal, so the model can better generalize the general patterns of normal samples, and then treats samples outside the normal patterns as abnormal samples. As shown in Fig. 6(d), for leather, although the distribution of real and simulated abnormal and normal samples in two-dimensional space is not ideal, MemSeg can finally complete the accurate defect localization at the pixel-level with an AUC score of 99.7%.

*4.5. Impact of Different Losses*

In MemSeg, we use L1 loss and focal loss as the loss functions. In Table 4, we report the AUC scores of MemSeg for anomaly localization when using different loss functions.

**Table 4.** The AUC scores of MemSeg on MVTec AD dataset when using different loss functions

| L1 Loss | Focal Loss [27] | Image-level | Pixel-level |
|---|---|---|---|
| √ |   | 84.82 | 73.38 |
|   | √ | 98.92 | 98.64 |
| √ | √ | **99.56** | **98.84** |

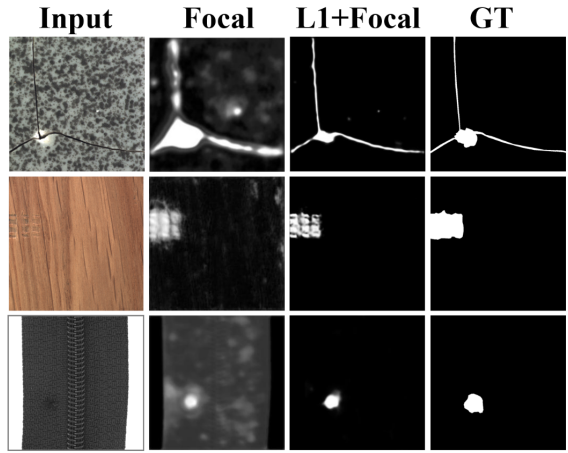

Fig.7. The effect of L1 loss on anomaly localization. When L1 loss and focal loss are used simultaneously, the edges of the segmented images obtained by MemSeg are more precise.

Since the ratio of normal samples to simulated abnormal samples is 1:1 in the training phase, and the proportion of abnormal regions in the simulated abnormal samples to the image is small, there exists data imbalance in the training samples at the pixel-level, so the model is difficult to train using only L1 loss. Therefore, focal loss needs to be introduced to make the model focus more on the abnormal regions in the training samples. As shown in Fig. 7, the simultaneous use of L1 loss and focal loss helps the model to learn better anomaly discrimination while converging more quickly.

*4.6. Impact of Module Components*

This subsection mainly discusses the effects of the memory module, multi-scale strategy, spatial attention, and CA [26] on the model. As shown in Table 5, memory information has a significant effect on abnormal localization. To further explore the effect of memory sample size on anomaly localization, we report the changes of AUC scores when the number of memory samples is 1, 15, 30, and 70 in Fig. 8. Within a certain range, as the number of memory samples increases, the model locates the abnormal regions more accurately, but when the number of memory samples is too large, it causes insufficient training samples and leads to a slight decrease on AUC scores, so an appropriate number of memory samples is crucial for MemSeg.

**Table 5.** The AUC scores of MemSeg on the MVTec AD dataset when using different module components.

| Memory | Multi-scale | Spatial Attention | Coordinate Attention | Image-level | Pixel-level |
|---|---|---|---|---|---|
|  |  |  |  | 96.42 | 96.08 |
| √ |  |  |  | 98.41 | 98.27 |
| √ | √ |  | √ | 99.08 | 98.60 |
| √ | √ | √ |  | 99.26 | 98.67 |
| √ |  | √ | √ | 98.96 | 98.44 |
| √ | √ | √ | √ | **99.56** | **98.84** |

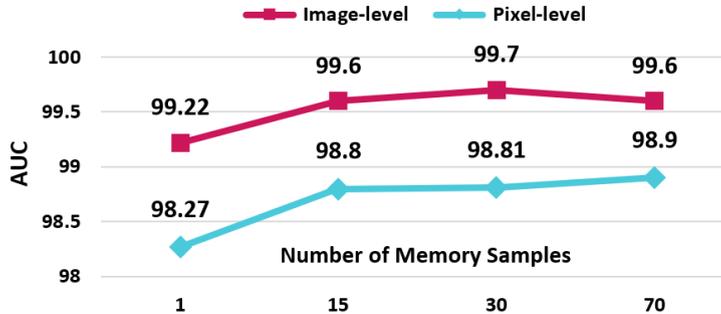

Fig.8. Effects of the different number of memory samples on AUC scores. The vertical coordinates report the mean AUC scores of the 13 categories in the MVTec AD dataset, excluding screw and toothbrush.

As shown in Table 5, the multi-scale strategy, spatial attention, and CA contribute significantly to the anomaly detection performance, and the multi-scale strategy has the greatest impact. In Fig. 9, we visualize the generation process of spatial attention maps $M_1$, $M_2$ and $M_3$ following the steps in Subsection 3.2. As shown in the figure, the best difference information $DI_n^*$ ($n = 1,2,3$) with different scales, which is generated relying on the memory information, already contains a blurred guess of the abnormal region in the input image. In the process of generating the spatial attention maps from $DI_n^*$, after multi-scale fusion, the noise in the heat map becomes less and the guess of the abnormal regions in the image becomes more certain. Although the final generated spatial attention map $M_1$ is still different from the final prediction of MemSeg to some extent, it is close to the ground truth of the abnormal regions, which visually demonstrates the effectiveness of memory information, multi-scale fusion, and spatial attention.

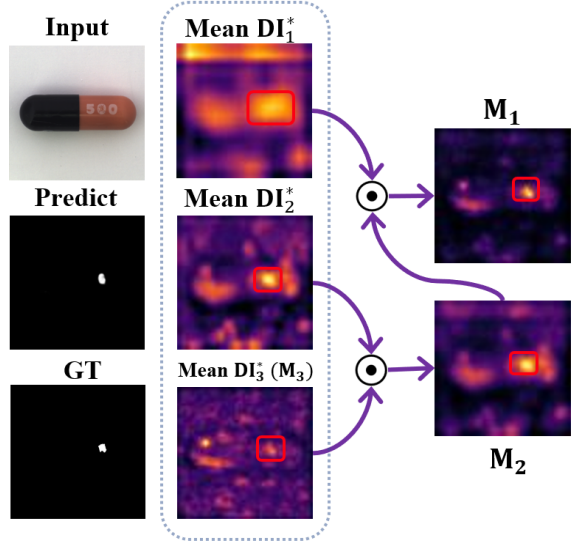

Fig.9. The generation process of spatial attention maps $M_1$, $M_2$ and $M_3$. This process visually demonstrates the effectiveness of memory module, multi-scale strategy, and spatial attention for image defect localization.

*4.7. Evaluation with A Toy Dataset*

The noise generated by MemSeg using the anomaly simulation strategy is irregular. To verify the ability of MemSeg to generalize regular noise, we generate a toy dataset using normal samples in the test set of MVTec AD dataset. As shown in Fig. 10, the shapes of the generated noise are rectangle, triangle, lightning bolt, star, heart, and circle; and the size, color, position, angle, and aspect ratio of the noise are random. For the abnormal samples in the toy dataset, it is never seen in the training phase of MemSeg. We apply the trained model directly to the toy dataset and compare the performance with three models. The AUC scores of the four models are shown in Table 6. MemSeg achieves precise localization of the abnormal regions with an AUC score close to 100%, which further demonstrates the strong generalization ability of our model to localize unknown anomalies.

Table 6. AUC scores of PatchCore, SPADE, PaDiM, and MemSeg on the toy dataset.

|  | PatchCore [11] | SPADE [12] | PaDiM [13] | Ours |
|---|---|---|---|---|
| Image-level | 99.75 | 95.75 | 99.30 | **99.83** |
| Pixel-level | 99.33 | 98.72 | 98.70 | **99.77** |

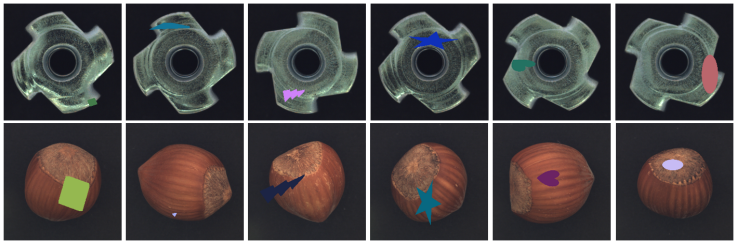

Fig.10. The toy dataset is generated using six regular shapes and regular colors to verify the generalization ability of MemSeg to regular noise.

*4.8. Inference Speed*

Compared to reconstruction-based methods [2,3,4,5,6,7], the embedding-based methods [10,11,12,13] achieve better performance in semi-supervised defect detection of images, but this kind of models needs to perform complex feature matching in the inference phase, which is difficult to be applied in industrial scenarios with high real-time requirements. Therefore, we are also interested in the inference speed of MemSeg. Our experiments are carried out on a PC with an NVIDIA RTX 3090 GPU. We calculate the time consumption of PaDiM [13] and SPADE [12] in the inference phase, and the time to process one image is 0.319s and 0.339s for these two models, respectively, while the time to process one image is 0.0319s for MemSeg, which is a 10-fold improvement in the inference speed. Meanwhile, compared with the reconstruction-based models, MemSeg avoids the conventional way of reconstructing the input samples and achieves the segmentation of abnormal regions in an end-to-end manner, which also has a competitive advantage in the inference speed.

## 5. Conclusion

Considering the small intra-class variation of products in the same production line, MemSeg simplifies the semi-supervised image surface defect detection into a simple and straightforward semantic segmentation task by introducing well-designed anomaly simulation strategy and memory information from the perspective of difference and commonality. Simple but high performance, MemSeg achieves SOTA performance while meeting the real-time requirements of industrial scenarios.

However, consistent with most embedding-based methods, the introduction of memory information makes MemSeg have some requirements on the spatial location of the detection targets in the images. When the location or direction of the detection targets is randomly distributed, such as screws in the MVTec AD dataset, MemSeg requires more memory samples to memorize more normal patterns, which places higher demands on the quality and quantity of the datasets.

**References**

[1] P. Bergmann, M. Fauser, D. Sattlegger, and C. Steger, "MVTec AD-a comprehensive real-world dataset for unsupervised anomaly detection," In: *Proceedings of the IEEE/CVF Conference on Computer Vision and Pattern Recognition*, pp. 9592-9600, 2019.
[2] P. Bergmann, S. Löwe, M. Fauser, D. Sattlegger, and C. Steger, "Improving unsupervised defect segmentation by applying structural similarity to autoencoders," arXiv preprint arXiv:1807.02011, 2018.
[3] T. W. Tang, W. H. Kuo, J. H. Lan, C. F. Ding, H. Hsu, and H. T. Young, "Anomaly detection neural network with dual auto-encoders GAN and its industrial inspection applications," *Sensors*, vol. 20, no. 12, pp. 3336, 2020.
[4] S. Akcay, A. Atapour-Abarghouei, and T. P. Breckon, "Ganomaly: Semi-supervised anomaly detection via adversarial training," In: *Asian conference on computer vision*, pp. 622-637, 2018.
[5] T. Schlegl, P. Seeböck, S. M. Waldstein, U. Schmidt-Erfurth, and G. Langs, "Unsupervised anomaly detection with generative adversarial networks to guide marker discovery," In: *International conference on information processing in medical imaging*, pp. 146-157, 2017.
[6] H. Zenati, C. S. Foo, B. Lecouat, G. Manek, and V. R. Chandrasekhar, "Efficient gan-based anomaly detection," arXiv preprint arXiv:1802.06222, 2018.
[7] T. Schlegl, P. Seebock, S. M. Waldstein, U. Schmidt-Erfurth, and G. Langs, "Unsupervised anomaly detection with generative adversarial networks to guide marker discovery," In: *International Conference on Information Processing in Medical Imaging*, pp. 146-157, 2017.
[8] J.B. Wheeler and H.A. Karimi, "A semantically driven self-supervised algorithm for detecting anomalies in image sets," *Computer Vision and Image Understanding*, vol. 213, 2021.
[9] L. C. Chen, G. Papandreou, I. Kokkinos, K. Murphy, and A. L. Yuille, "Deeplab: Semantic image segmentation with deep convolutional nets, atrous convolution, and fully connected crfs," *IEEE transactions on pattern analysis and machine intelligence*, vol. 40, no. 4, pp. 834-848, 2017.
[10] Y. Zheng, X. Wang, R. Deng, T. Bao, R. Zhao, and L. Wu, "Focus your distribution: coarse-to-fine non-contrastive learning for anomaly detection and localization," arXiv preprint arXiv:2110.04538, 2021.
[11] K. Roth, L. Pemula, J. Zepeda, B. Schölkopf, T. Brox and P. Gehler, "Towards total recall in industrial anomaly detection," arXiv preprint arXiv:2106.08265, 2021.
[12] N. Cohen and Y. Hoshen, "Sub-image anomaly detection with deep pyramid correspondences," arXiv preprint arXiv:2005.02357, 2020.
[13] T. Defard, A. Setkov, A. Loesch, and R. Audigier, "Padim: a patch distribution modeling framework for anomaly detection and localization," In: *International Conference on Pattern Recognition*, pp. 475-489, 2021.
[14] K. He, X. Zhang, S. Ren, and J. Sun, "Deep residual learning for image recognition," In: *Proceedings of the IEEE conference on computer vision and pattern recognition*, pp. 770-778, 2016.
[15] V. Zavrtanik, M. Krista, and D.Skočaj, "Reconstruction by inpainting for visual anomaly detection," *Pattern Recognition*, vol. 112, 2021.
[16] J. Pirnay and K.Chai, "Inpainting transformer for anomaly detection," arXiv preprint arXiv:2104.13897, 2021.

[17] Zavrtanik, Vitjan, M. Kristan, and D. Skocaj, "DRAEM-a discriminatively trained reconstruction embedding for surface anomaly detection," In: *Proceedings of the IEEE/CVF International Conference on Computer Vision*, pp. 8330-8339, 2021.
[18] C. L. Li, K. Sohn, J. Yoon, and T. Pfister, "CutPaste: self-supervised learning for anomaly detection and localization," In: *Proceedings of the IEEE/CVF Conference on Computer Vision and Pattern Recognition*, pp. 9664-9674, 2021.
[19] J. Song, K. Kong, Y. I. Park, S. G. Kim, and S. J. Kang, "AnoSeg: anomaly segmentation network using self-supervised learning," arXiv preprint arXiv:2110.03396, 2021.
[20] J. Deng, W. Dong, R. Socher, L. J. Li, K. Li, and L. Fei-Fei, "ImageNet: a large-scale hierarchical image database," In: *2009 IEEE conference on computer vision and pattern recognition*, pp. 248-255, 2009.
[21] O. Ronneberger, P. Fischer, and T. Brox, "U-net: convolutional networks for biomedical image segmentation," In: *International Conference on Medical image computing and computer-assisted intervention*, pp. 234-241, 2015.
[22] K. Perlin, "An image synthesizer," *ACM Siggraph Computer Graphics*, vol. 19, no. 3, pp. 287–296, 1985.
[23] M. Cimpoi, S. Maji, I. Kokkinos, S. Mohamed, and A. Vedaldi, "Describing textures in the wild," In: *Proceedings of the IEEE Conference on Computer Vision and Pattern Recognition*, pp. 3606–3613, 2014.
[24] T. Y. Lin, P. Dollár, R. Girshick, K. He, B. Hariharan, and S. Belongie, "Feature pyramid networks for object detection," In: *Proceedings of the IEEE conference on computer vision and pattern recognition*, pp. 2117-2125, 2017.
[25] S. Chen, Z. Cheng, L. Zhang, and Y. Zheng, "SnipeDet: Attention-guided pyramidal prediction kernels for generic object detection," *Pattern Recognition Letters*, vol. 152, pp. 302-310, 2021.
[26] Q. Hou, D. Zhou, and J. Feng, "Coordinate attention for efficient mobile network design," In: *Proceedings of the IEEE/CVF Conference on Computer Vision and Pattern Recognition*, pp. 13713-13722, 2021.
[27] T. Y. Lin, P. Goyal, R. Girshick, K. He, and P. Dollár, "Focal loss for dense object detection," In: *Proceedings of the IEEE international conference on computer vision*, pp. 2980-2988, 2017.
[28] P. Mishra, R. Verk, D. Fornasier, C. Piciarelli, and G. L. Foresti, "VT-ADL: a vision transformer network for image anomaly detection and localization," arXiv preprint arXiv:2104.10036, 2021.
[29] J. Yi and S. Yoon, "Patch svdd: patch-level svdd for anomaly detection and segmentation," In: *Proceedings of the Asian Conference on Computer Vision*, 2020.
[30] J. Y. Ahn and G. Kim, "Application of optimal clustering and metric learning to patch-based anomaly detection," *Pattern Recognition Letters*, 2022.
[31] L. Van der Maaten and G. Hinton, "Visualizing data using t-SNE," *Journal of machine learning research*, vol. 9, no. 11, 2008.